\documentclass[10pt,twocolumn,letterpaper]{article}

\usepackage{cvpr}
\usepackage{times}
\usepackage{epsfig}
\usepackage{graphicx}
\usepackage{amsmath}
\usepackage{amssymb}

\DeclareMathOperator*{\argmin}{argmin} 
\usepackage{booktabs}
\usepackage{xcolor,colortbl}
\usepackage{tabularx}
\usepackage{multirow}
\usepackage{paralist}

\usepackage[pagebackref=true,breaklinks=true,letterpaper=true,colorlinks,bookmarks=false]{hyperref}

\cvprfinalcopy 

\def\httilde{\mbox{\tt\raisebox{-.5ex}{\symbol{126}}}}

\begin{document}

\title{Learning Anthropometry from Rendered Humans}

\author{Song Yan and Joni-Kristian Kämäräinen\\
Computer Sciences, Tampere University, Finland\\
{\tt\small firstname.lastname@tuni.fi}
}

\maketitle\def\httilde{\mbox{\tt\raisebox{-.5ex}{\symbol{126}}}}

\begin{abstract}
  Accurate estimation of anthropometric body measurements from RGB imags
  has many potential applications in industrial design, online clothing,
  medical diagnosis and ergonomics.
  Research on this topic is limited by the fact that there exist only
  generated datasets which are based on fitting a 3D body mesh to
  3D body scans in the commercial CAESAR dataset. For 2D only silhouettes
  are generated.
  To circumvent the data bottleneck, we introduce a new 3D scan dataset of 2,675 female and 1,474 male scans.
  We also introduce a small dataset of 200 RGB images and tape measured ground truth.
  With the help of the two new datasets we propose a part-based shape model and a deep neural network for estimating anthropometric measurements from 2D images. All data will be made publicly available.

\end{abstract}


\section{Introduction}
\label{sec:intro}
Recovery of 3D human body from 2D images is an important yet challenging problem with many potential applications in industrial design~\cite{Park-2017-injury}, online clothing~\cite{Daanen-2008-IJCST}, medical diagnosis~\cite{Ogden-report} and work ergonomics~\cite{Pheasant-bodyspace-book}. 
However, compared to pose estimation, less attention has been paid on the task of accurate shape estimation, especially from RGB images. 
Due to the lack of public datasets, previous works~\cite{dibra2016shape, dibra2016hs, Dibra-2017-cvpr,boisvert2013three, chen2011silhouette, chen2010inferring, chen2009learning, sigal2008combined, xi2007data} adopt the strategy of creating synthetic samples with shape models, e.g. SMPL~\cite{SMPL:2015} and SCAPE~\cite{Anguelov-2005-siggraph}, and reconstruct body shapes from generated 2D silhouettes.
Recent works~\cite{guan2009estimating, Bogo:ECCV:2016, Unite-the-People, kolotouros2019spin, kanazawa2018end, pavlakos2018learning} consider to directly estimate human bodies from RGB images, but the works focus
on 3D pose estimation.

Vision based anthropometry has many potential applications in clothing industry, custom tailoring, virtual fitting and games.
The state-of-the-art works~\cite{dibra2016shape, dibra2016hs, Dibra-2017-cvpr} recover 3D body surfaces from silhouettes and obtain the anthropometric measurements as by-products. There does not exist
an RGB dataset for evaluation and HS-Net in~\cite{dibra2016hs} and HKS-Net in~\cite{Dibra-2017-cvpr} are evaluated only on 4-7 real samples.

To tackle the task of accurate anthropometric measurement estimation from RGB images, we directly regress 2D images to body measurements using a
deep network architecture which omits the body reconstruction stage.
However, we also provide a 3D body mesh by learning a mapping from the
measurements to the shape coefficients of a part-based shape model.
For network training and shape model building, we introduce a new dataset
of 3D body scans. For training we render virtual RGB bodies consistent
with the true data.
To evaluate measurement prediction for real cases, we also release a testing RGB dataset of 200 real subjects and their tape measurements as ground truth. The proposed network, trained with generated data, provide
anthropometric measurements with state-of-the-art accuracy as compared
to the previous works on the existing~\cite{yang2014semantic, pishchulin17pr} and the new introduced data.
\begin{figure}[t]
    \centering
    \includegraphics[width=0.85\linewidth]{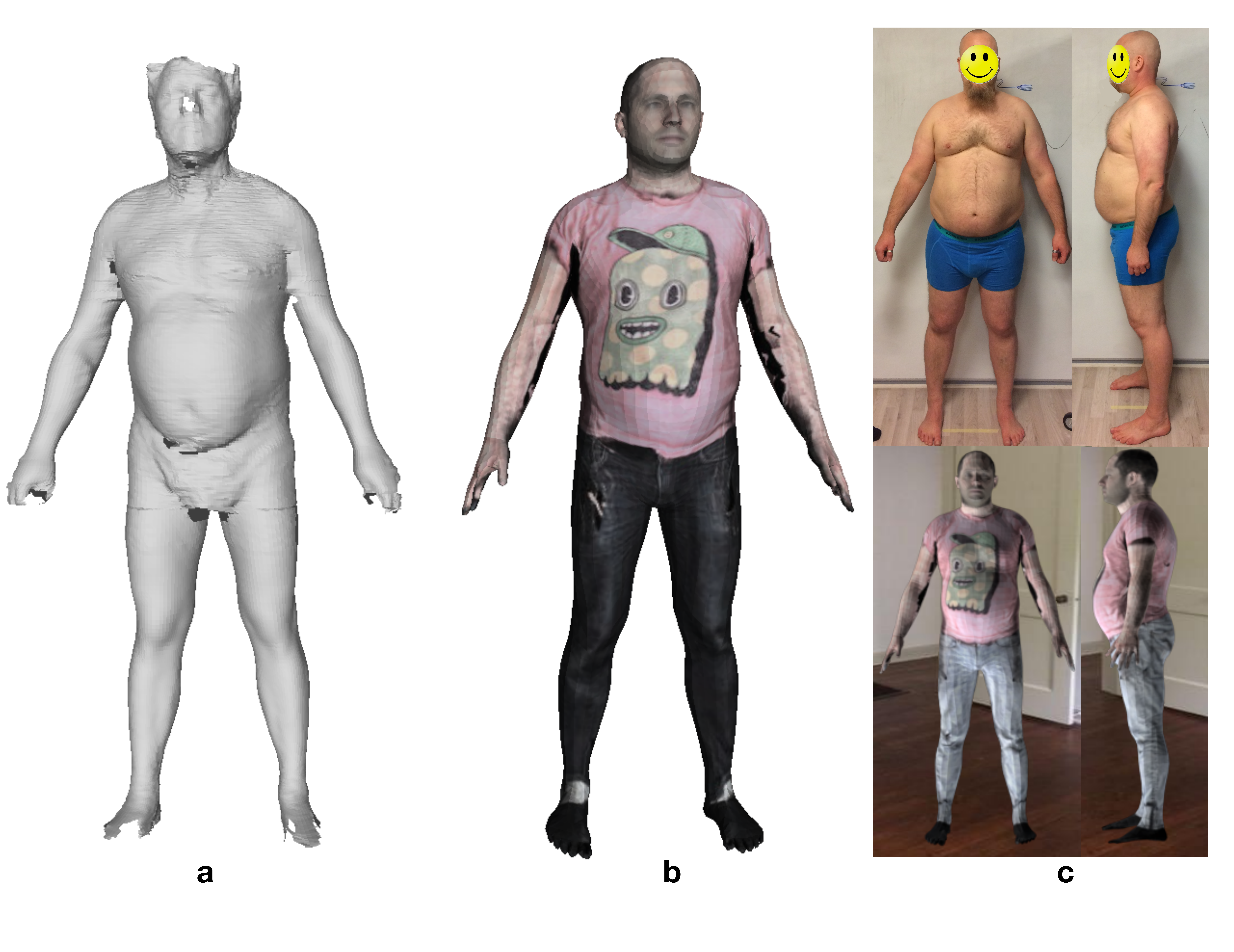}
   \caption{Contributions of this work: a) 4,149 3D body scans; b) fitted body meshes with textures; and c) real (top) and rendered (bottom) RGB images for training anthropometric measurement network architectures.}
    \label{fig:dataset}
\end{figure}

{\bf{Contributions}} of our work are the following:
\begin{compactitem}
\item[-] a dataset of 2,675 female and 1,474 male scans,
\item[-] A dataset of 200 RGB images of real subjects with tape measured ground truth;
\item[-] an anthropometric body measurement network architecture trained with rendered images.
\end{compactitem}
In the experiments our network
achieves competitive performances on both tasks of anthropometric measurement estimation and body shape reconstruction compared to the state-of-the-art works.

\section{Related work}
\label{sec:related}

\noindent{\bf{Datasets.}} 
CAESAR dataset~\cite{robinette2002civilian} is a commercial dataset of human body scans and tape measured anthropometric measurements and its lisence prevents public usage.
Yang~\etal~\cite{yang2014semantic} and Pischulin~\etal~\cite{pishchulin17pr}
fitted 3D body templates to the CAESAR scans and used the fitted meshes
and geodesic distances in their experiments. Some of the fitted meshes
are available in their project pages.
Another popular option has been to use synthetic data, for example,
SURREAL~\cite{varol2017learning} consists of synthetic 3D meshes and RGB images rendered from 3D sequences of human motion capture data and by
fitting the SMPL body model~\cite{SMPL:2015}. Realistic dataset with
RGB images and tape measured ground truth is not available. In this work we use the
fitted CAESAR meshes~\cite{yang2014semantic}, namely CAESAR fits.

\noindent{\bf{Shape models.}} 
The body shape variation is typically captured by principal component
analysis (PCA) on registered meshes of a number of subjects, such as the
3D scans in the commercial CAESAR dataset~\cite{robinette2002civilian}.
For example, Allen et al~\cite{allen2003space} fit a template to a subset of CAESAR dataset and model the body shape variation by PCA.
Seo et al.~\cite{Seo-2003-siggraphsymp} adopt the same approach in
their characterization of body shapes.
SCAPE~\cite{Anguelov-2005-siggraph} is one of the most popular shape models 
used in similar works to ours. SCAPE decomposes shape to pose invariant
and pose dependent shape components to perform more realistic
deformations.
%
Yang et.al~\cite{yang2014semantic} utilize the SCAPE model to learn the shape deformation and introduce a local mapping method from anthropometric measurements ("semantic parameters") to shape deformation parameters.
Another popular shape model is SMPL~\cite{SMPL:2015} which also decomposes shape into pose dependent and independent parts. SMPL shape
variation is also learned from the CAESAR data, but provides better
details than SCAPE. 
The public version of the SMPL model
provides only 10 PCA components preventing reconstruction of local details.

\begin{figure}[t]
    \centering
    \includegraphics[width=0.9\linewidth]{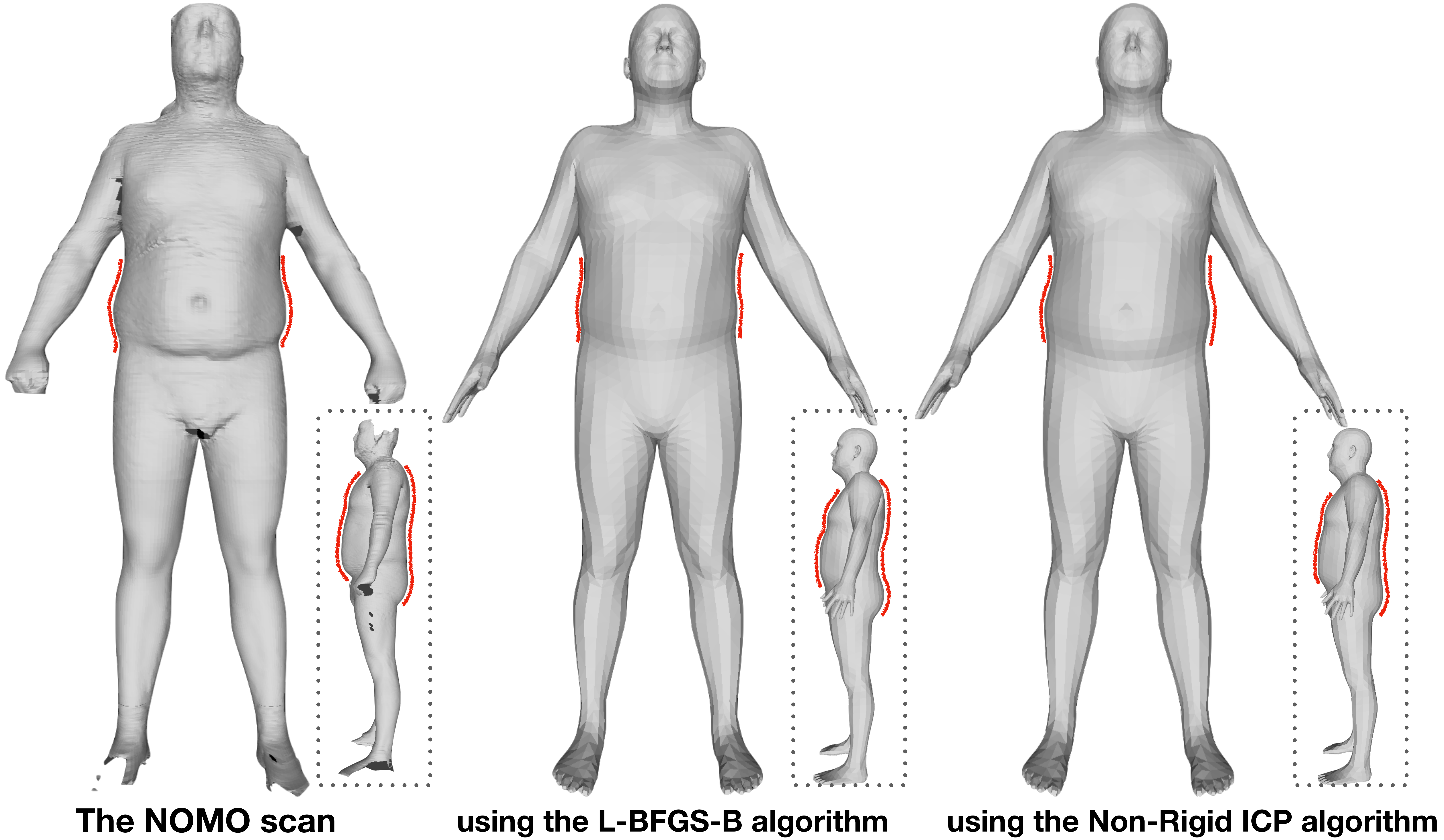}
   \caption{Left: an original scan of a subject. 
   Middle: SMPL fit using the L-BFGS-B algorithm~\cite{zhu1997algorithm}.
   Right: the registered mesh using Non-Rigid ICP algorithm~\cite{amberg2007optimal}. 
   Red curves mark the local areas where the two methods particularly differ from each other.}
    \label{fig:nicp_fits}
\end{figure}

One drawback of PCA based shape modelling is the fact that PCA vectors
represent global deformation and important details of local parts such
as upper torso or pelvis can be missing (Figure~\ref{fig:nicp_fits}).
There exists a number of shape models that provide local deformations.
For example, Zuffi et al.~\cite{Zuffi-2015-cvpr} introduce a part-based model in which each body part can independently deform.
Similarly Bernard~\etal and Neumann~\etal~\cite{bernard2016linear, neumann2013sparse} extract sparse and spatially localized deformation factors for better local shape deformation.

Auxiliary information, such as the qualitative body type, has been added to the shape
parameters in several
works~\cite{allen2003space,streuber2016body, seo2003synthesizing,Seo-2003-siggraphsymp,yang2014semantic}.

\noindent{\bf{Shape estimation}} 
Due to the lack of real RGB datasets, previous works~\cite{boisvert2013three, chen2011silhouette, chen2010inferring, chen2009learning, sigal2008combined, xi2007data, Dibra-2017-cvpr, dibra2016hs, dibra2016shape, ji2018shape, seo20063d} reconstruct 3D body meshes from 2D silhouettes.
The silhouettes are generated using the CAESAR fits or using synthetic body models.
The early works extract handcrafted silhouette features which are mapped to 3D shape
parameters using, e.g., the linear mapping~\cite{xi2007data}, a mixture of kernel regressors~\cite{sigal2008combined}, Random Forest Regressors~\cite{dibra2016shape,chen2011silhouette}, or a shared Gaussian process latent variable model~\cite{chen2010inferring,chen2009learning}.
The more recent works~\cite{dibra2016hs,Dibra-2017-cvpr,ji2018shape} propose deep network
architectures to estimate the shape model parameters in an end-to-end manner.

A number of pose estimation methods also provide a 3D shape estimate~\cite{guan2009estimating, Bogo:ECCV:2016, Unite-the-People, kolotouros2019spin, kanazawa2018end, pavlakos2018learning}, but
shape is only coarse and anthropometric measurements made on them are inaccurate (see our experiments).
In these works, a parametric 3D model is fitted to silhouettes~\cite{hasler2010multilinear},
certain body keypoints or joints~\cite{Bogo:ECCV:2016}, or a collection of 2D observations~\cite{guan2009estimating,Unite-the-People}.
For example, given the subject's height and a few clicked points, Guan et
al.~\cite{guan2009estimating} fits the SCAPE model to the image and
fine-tuners the result using silhouettes, edges and shadings.
Kanazawa et al.~\cite{kanazawa2018end} propose an end-to-end adversarial learning framework to
recover 3D joints and body shapes from a single RGB image by minimizing the joint reprojection error.
Kolotouros et al.~\cite{pavlakos2018learning} extend SMPLify~\cite{Bogo:ECCV:2016}
by neural network based parameter initialisation and iterative optimization.
To estimate 3D human shapes from measurements,~\cite{wuhrer2013estimating} first
optimize the shape of a PCA-based model to find the landmarks that best describe
target measurements and then deform the shape to fit the measurements.

\noindent{\bf{Anthropometric measurements.}}
Previous works~\cite{Tsoli-2014-wacv, weiss2011home, markiewicz20173d, wasenmuller2015precise, yan2019anthropometric} predict measurements from 3D human scans with the help of 3D body models which provide the correspondences.
They first register a template to scans, then obtain the lengths of measurement paths defined by the vertices on the template surface (geodesic distances).
From registered meshes, Tsoli et al.~\cite{Tsoli-2014-wacv} extract the global and local features, including PCA coefficients of triangle deformations and edge lengths, the circumferences and limb lengths, then predicts measurements from these features using regularized linear regression. 
To eliminate negative effects caused by varying positions of measurement paths across subjects, \cite{yan2019anthropometric} obtains the optimal result through a non-linear regressor over candidate measurements extracted from several paths in the same area.

There exists a few works estimating anthropometric measurements from 2D images.
Most works~\cite{wang2003virtual, boisvert2013three, dibra2016hs, Dibra-2017-cvpr, lin2012constructing} first construct a shape model and then obtain measurements from reconstructed bodies.
Another line of works~\cite{aslam2017automatic, gordon20142012, wang2019new, lin2008automatic}
estimate the circumferences of body parts using fiducial points on the contours.
For example, in \cite{aslam2017automatic} part circumferences are estimated using an ellipsoid model and lengths between two relevant fiducial points from the frontal and lateral views of silhouettes.

To the authors' best knowledge our work is the first attempt to estimate accurate anthropometric measurements from RGB images.
%
\section{Methodology}
\label{sec:method}

\subsection{Datasets}
\label{sec:dataset}
\noindent{\bf{Rendered RGB.}}
We collected a dataset of real body scans, namely {\textit{XXXX-scans}} dataset,
captured by a commercial TC2 system\footnote{https://www.tc2.com}.
1,474 male and 2,675 female subjects were scanned.
The scanned subjects were instructed to take approximate “A”-pose and held
the capturing device handles. The quality of the scans vary and in many of the
scans point regions are missing near feet, hand and head. To construct
watertight meshes an SMPL template in the "A"-pose was fitted using the non-rigid
ICP method of Amberg~\etal~\cite{amberg2007optimal} (Fig~\ref{fig:dataset} a-b \& Fig~\ref{fig:nicp_fits} Right).
The fitted XXXX-scans dataset is called as {\textit{XXXX-fits}}.

Finally, a set of {\em rendered RGB images} were generated from the XXXX-fits meshes
using the rendering method in SURREAL~\cite{varol2017learning}. Each image was generated
using a randomly selected home background, body texture (clothes and skin details),
lighting and a fixed camera position. RGB images were generated from the both
frontal and lateral views (Figure~\ref{fig:dataset} c (bottom)).

\noindent{\bf{Real RGB.}}
We collected a dataset of RGB images of 200 volunteers using iPhone 5S rear camera (Figure~\ref{fig:dataset} c (top)), namely {\textit{XXXX-real-200}}.
All volunteers wear only underwear and photos were captured indoors.
The approximate capturing distance was 2.4~m and the camera height from the ground
1.6~m. The anthropometric measurements were done using a tape measure by a
tailoring expert.

\subsection{Part-based Shape Model}
\label{sec:subpcashapemodel}
\begin{figure}[t]
    \centering
    \includegraphics[width=0.95\linewidth]{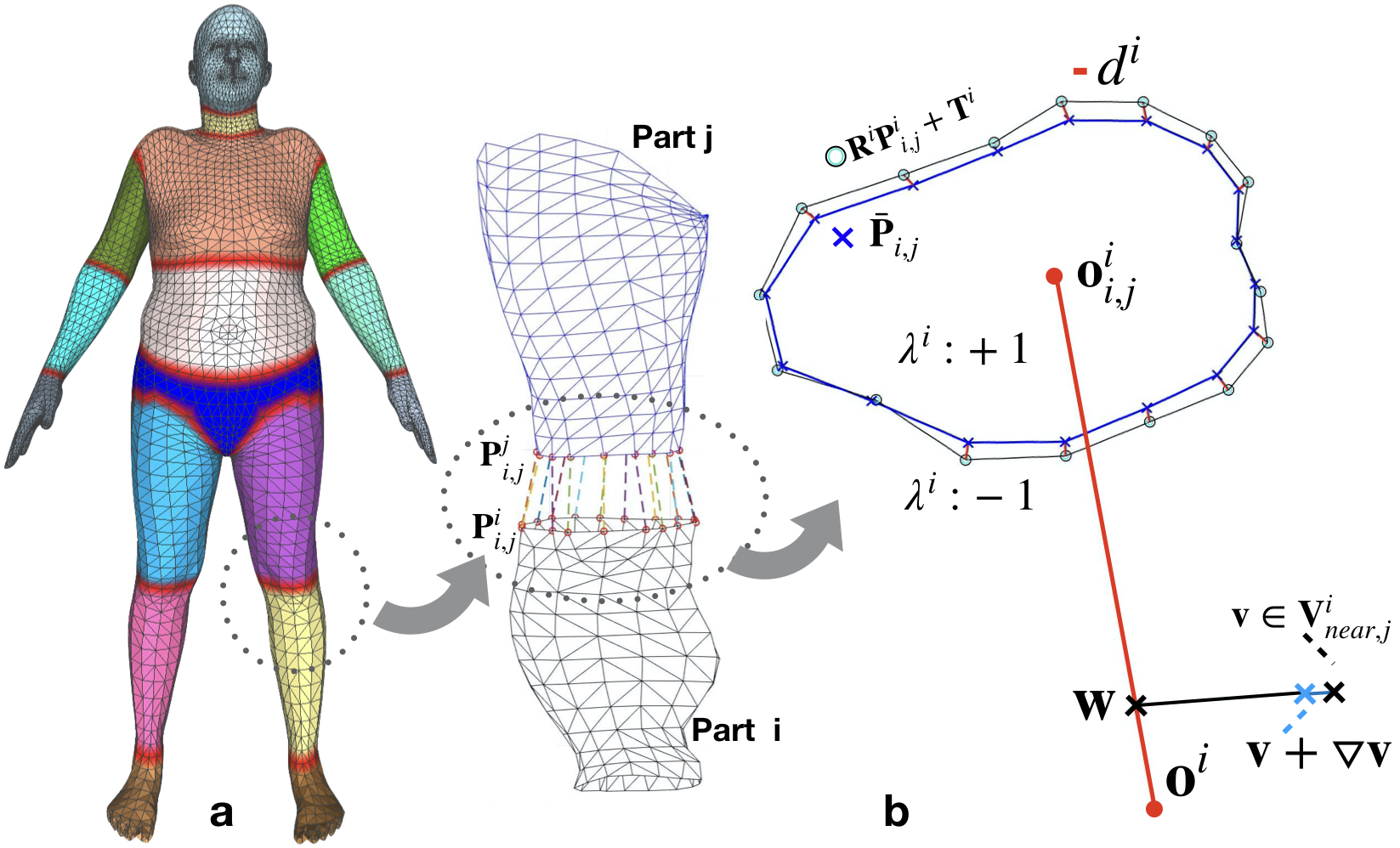}
   \caption{The Body Segments (a) and the Stitching Process (b). Details of the stitch deformation are illustrated in Sec~\ref{sec:subpcashapemodel}. Red markers in (a) denote the interface points.}
    \label{fig:stitchprocess}
\end{figure}
Since in our scenario the subject is volunteering and takes
a pre-defined pose it can be safely assumed that decomposition of
the shape to pose specific and pose invariant components is not needed.
To capture local details, we adopt a part-based body model to be able to model shape
variation of each part with the same accuracy.
The proposed model is composed of 17 body parts: head, neck, upper torso, lower torso, pelvis, upper legs, lower legs, upper arms, lower arms, hands and feet (Figure~\ref{fig:stitchprocess} a).
Each part is a triangulated 3D mesh in a canonical, part-centered, coordinate system.

\paragraph{Part-based Shape Model.}
The SP model~\cite{Zuffi-2015-cvpr} first applies PCA over the full body and then defines a PCA component matrix for each part by grouping part specific rows in the shape basis matrix. Instead we directly apply PCA on each body part to model its shape variance.
Let $\bold{X}^i$ be the mesh vertices for the part $i$, a part instance is generated by adjusting the shape parameters $\bold{\beta}^i$ as
\begin{equation}
    \label{eq:shapemodelequation}
    \bold{X}^i = \bold{U}^i \bold{\beta}^i + \bold{\mu}^i
\end{equation}
where $\bold{U}^i$ is the PCA component matrix and $\bold{\mu}^i$ is the mean intrinsic shape across all training shapes.

\paragraph{Part Stitching.}
Inspired by~\cite{Zuffi-2015-cvpr}, we also define interface points $\bold{P}_{i,j}$ that are shared by two neighbor parts $i$ and $j$.
The stitching process (see Figure~\ref{fig:stitchprocess} b) starts from the root node (the pelvis) and stitches the part $i$ with its parent $j$ using the rotation matrix $\bold{R}^i$ and translation matrix $\bold{T}^i$.
Translation and rotation are solved as the Orthogonal Procrustes transformation:
\begin{align}
\begin{split}
    \label{eq:stitchequation}
    \bold{R}^i ~~&= \argmin_\Omega ||\bold{\Omega} \bold{P}^i_{i,j} - \bold{P}^j_{i,j}||_F, ~s.t.~ \Omega^T\Omega = I \\
    \bold{T}^i ~~&= \bold{o}^j_{i,j} - \bold{o}^i_{i, j} 
\end{split} 
\end{align}
where $\bold{P}^i_{i,j}$, $\bold{P}^j_{i,j}$ denote the interface points on the part $i$ and $j$ respectively, 
and $\bold{o}^i_{i,j}$, $\bold{o}^j_{i, j}$ indicate the centers of the interface points 
and $\bold{X}^{i,align}$, $\bold{P}^{i, align}_{i,j}$ represent the aligned mesh vertices and the interface points of part $i$.
We adopt $\bar{\bold{P}}_{i, j} = (\bold{P}^{i, align}_{i,j} + \bold{P}^j_{i,j})/2$ as the final interface points.

Neighbor parts of the same body should be stitched seamlessly.
Hence we introduce the stitch deformation to smooth the stitched areas.
Consider the part $i$ as the example, 
we calculate the mean deformation distance $\triangledown d^i$ and the deformation direction $\lambda^i$ as follows:
\begin{align}
\begin{split}
    \label{eq:stitchDeformDist}
    \triangledown d^i &= \frac{1}{|\bar{\bold{P}}_{i, j}|} \sum^{|\bar{\bold{P}}_{i, j}|}_{k=1} dist(\bold{\bar{p}}^i_k, \bold{p}^{i,align}_k)\\
    \lambda^i &= \bigg\{ \begin{array}{l}
                        -1, ~~if~ \frac{1}{N} \sum^{N}_{k=1}
                        \measuredangle \bold{o}^i_{i,j} \bold{\bar{p}}^i_k \bold{p}^{i,align}_k \leq \frac{\pi}{2}
                        \\
                        +1, ~~otherwise
                        \end{array}
\end{split}
\end{align}
where $\bold{\bar{p}}^i_k$, $\bold{p}^{i, align}_k$ are the k-th points of $\bar{\bold{P}}_{i, j}$ and $\bold{P}^{i, align}_{i,j}$.
$\lambda^i$ indicates the deformations towards inside or outside. 

Let $\bold{o}^i$ be the center of part $i$, 
$\bold{v} \in \bold{V}^i_{near, j}$ be a random vertex near by the interface area,
and $\bold{w}$ be a point on the line segment $\bold{o}^i \bold{o}^i_{i,j}$
and ${\overrightarrow{\bold{v}\bold{w}}} \perp {\small{{\overrightarrow{\bold{o}^i \bold{o}^i_{i,j}}}}}$.
The deformation $\triangledown \bold{v}$ at vertex $\bold{v}$ can be presented as:
\begin{align}
\label{eq:stitchDeform}
\begin{split}
    \triangledown \bold{v} &= \lambda^i * \frac{\overrightarrow{\bold{v}\bold{w}}}{|\overrightarrow{\bold{v}\bold{w}}|}*\frac{|\overrightarrow{\bold{o}^i_{i,j}\bold{w}}|}{|\overrightarrow{\bold{o}^i\bold{o}^i_{i,j}}|} * \triangledown d^i\\
    \bold{V}^i_{near,j} &= \{\bold{v} | \frac{|\overrightarrow{\bold{o}^i_{i,j}\bold{w}}|}{|\overrightarrow{\bold{o}^i\bold{o}^i_{i,j}}|} \leq \epsilon, for~ \bold{v} \in \bold{X}_i\}
\end{split}
\end{align}
where $\bold{V}^i_{near, j}$ denotes the neighbour vertices of the interface points $\bold{P}^i_{i,j}$, and $\epsilon$ is set to 0.1 in our experiments.

\subsection{Virtual Tailor Body Measurements}
\begin{figure}
    \centering
    \includegraphics[width=0.95\linewidth]{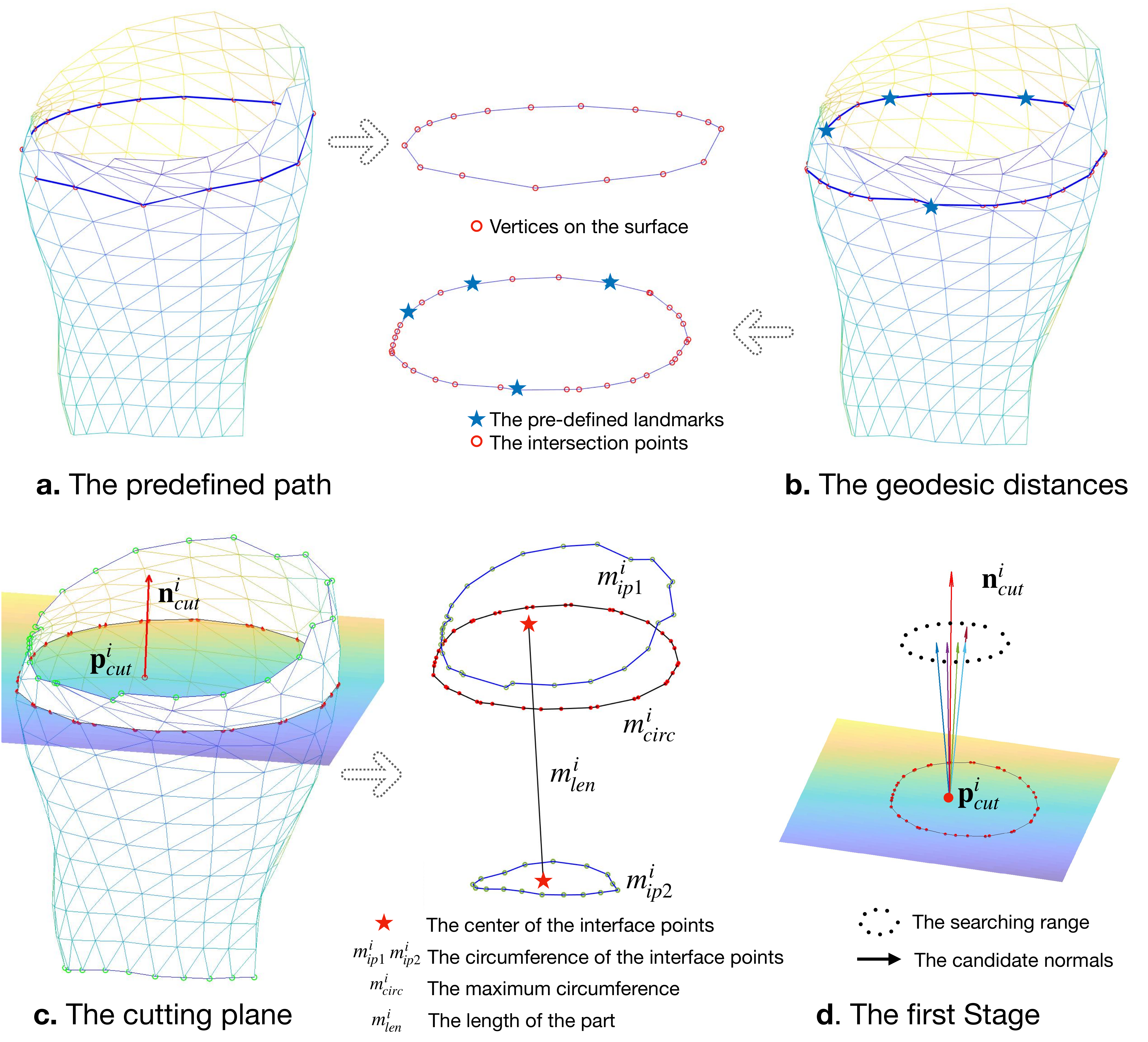}
    \caption{Various types of virtual tailor body measurements:
    a) a predefined path consisting of a set of vertices; 
    b) a geodesic path through a set of pre-defined landmarks;
    c) a perimeter of the plane section (proposed);
    d) the first stage of the proposed procedure finds the normal of the cutting plane.}
    \label{fig:measurementdefinition}
\end{figure}
\label{sec:measurementdefinition}
Accurate anthropometric body measurements are the final goal of
vision based body shape analysis. Therefore it is important how these
are defined when trained with 3D model rendered images. In prior
arts there have been two dominating practices (Figure~\ref{fig:measurementdefinition}):
i) predefined paths consisting of a set of vertices~\cite{yan2019anthropometric}; 
ii) geodesic distances through pre-defined landmarks~\cite{boisvert2013three,dibra2016hs, Dibra-2017-cvpr}.
The first method sums edge lengths between the pre-defined vertices. However,
due to the non-rigid ICP model fitting procedure the vertex positions can be heavily deformed and the paths do not anymore correspond to the
shortest path used by the tailors.
The second method defines a number of landmarks along
the circumference paths, but also the landmarks suffer from fitting deformations.
In order to provide measurements that better match the tailor procedure,
we propose an alternative measurements. Our procedure first aligns the body
mesh rotation and then uses a cutting plane to define a circumference path without deformations.

The perimeter of the surface along the plane section of each
body part $i$ is adopted as the circumference measure $m^i_{circ}$
of that part.
The cutting plane is determined by the cutting point $\bold{p}^i_{cut}$ and the normal $\bold{n}^i_{cut}$.
The whole process (Figure~\ref{fig:measurementdefinition} c-d) consists of two stages. 
The first stage finds the normal $\bold{n}^i_{cut}$ which minimizes the circumference $m^i_{circ}$ at the certain cutting point. 
This stage forces the cutting plane to be perpendicular to the body part. Finding the cutting point $\bold{p}^i_{cut}$ which maximizes the circumference $m^i_{circ}$ is done by sampling in a certain range in the second stage.

Besides the circumference $m^i_{circ}$, we also define the length $m^i_{len}$ of the body part and the circumferences $m^i_{ip1}, m^i_{ip2}, ... m^i_{ipN}$ of the interface points.
The measurements of the body part $i$ are $\bold{m}^i = [m^i_{circ}, m^i_{len}, m^i_{ip1}, m^i_{ip2}, ... m^i_{ipN}]^T$

\subsection{From Body Measurements to Body Shape}
\label{sec:semanticmapping}
\begin{figure}[t]
    \centering
    \includegraphics[width=0.95\linewidth]{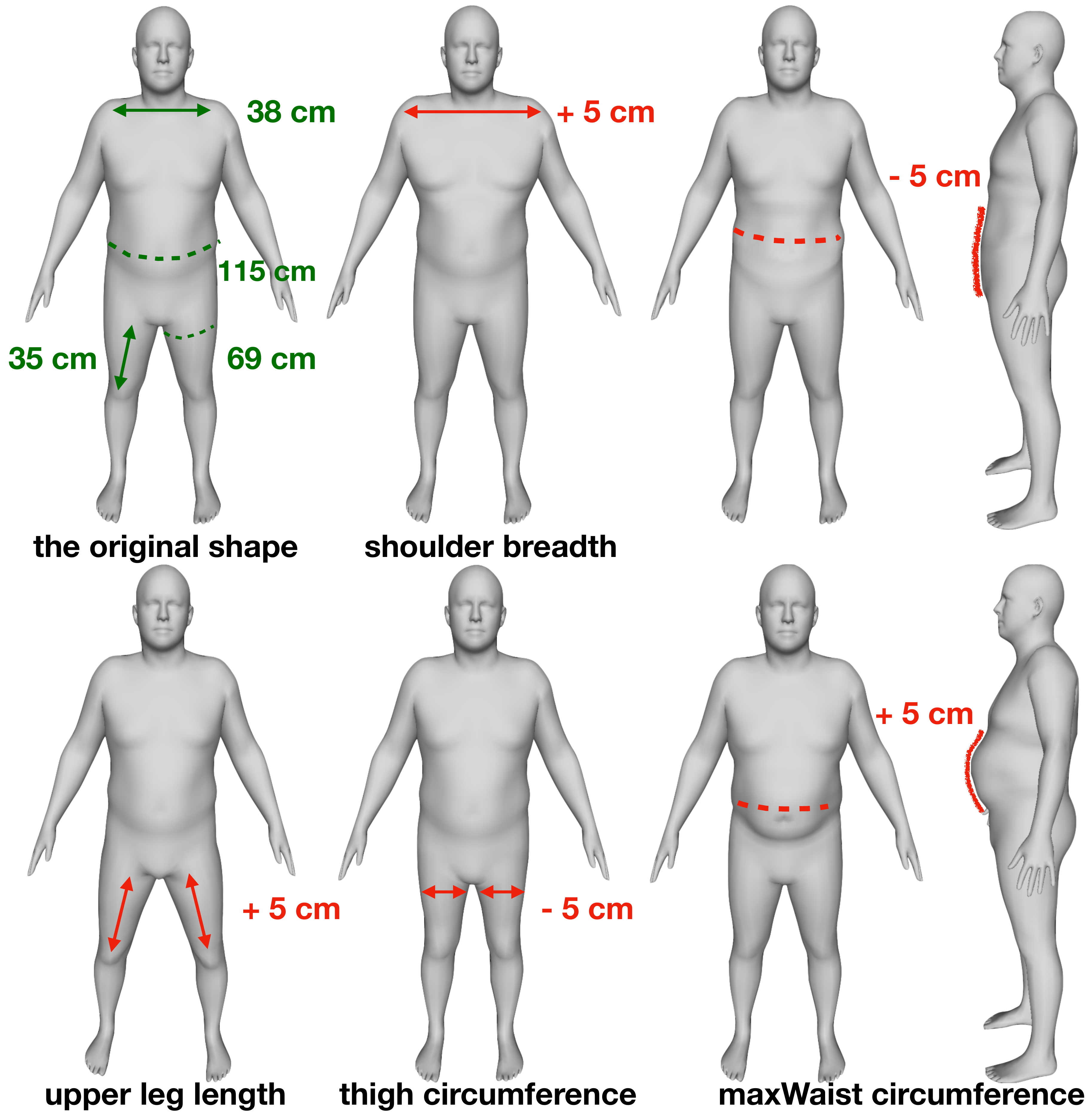}
   \caption{Editing the shape by updating the body measurements.
   {\bf{Green numbers}} denote the original body measurements. 
   {\bf{Red numbers}} denote increasing or decreasing these measurements.}
    \label{fig:editBody}
\end{figure}
Similar to~\cite{allen2003space, yang2014semantic}, we learn mapping
from the body measurements $\bold{m}^i$ to the PCA shape variables $\bold{\beta}^i$. This is done separately for each body part
using
\begin{align}
    \label{eq:mappingequation}
    [m^i_{circ}, m^i_{len}, m^i_{ip1}, ... m^i_{ipN}, 1]^T \bold{F}^i = \bold{\beta}^i
\end{align}
Using the training set, the computed measurements and shape parameters
are put into data matrices $\bold{M}^i$ and $\bold{B}^i$ and the transformation matrix $\bold{F}^i$ is computed in the least-square
sense as
\begin{align}
    \label{eq:leastsquare}
    \bold{F}^i = \bold{M}^{i, +} \bold{B}^i
\end{align}
where $^+$ denotes the pseudo-inverse operation.
Given a new set of body measurements $\bold{m}$, we can obtain the PCA coefficients from $\bold{b} = \bold{m}\bold{F}$.

Finally, mapping from the anthropometric measurements to body shapes
allows more intuitive generation of training samples as the measurements
can be varied $\pm \Delta \bold{m}$ and the corresponding body shapes
generated (Figure~\ref{fig:editBody}).

\subsection{Anthropometric Measurements Network}
\label{sec:measnetwork}
\begin{figure}[t]
    \centering
    \includegraphics[width=0.95\linewidth]{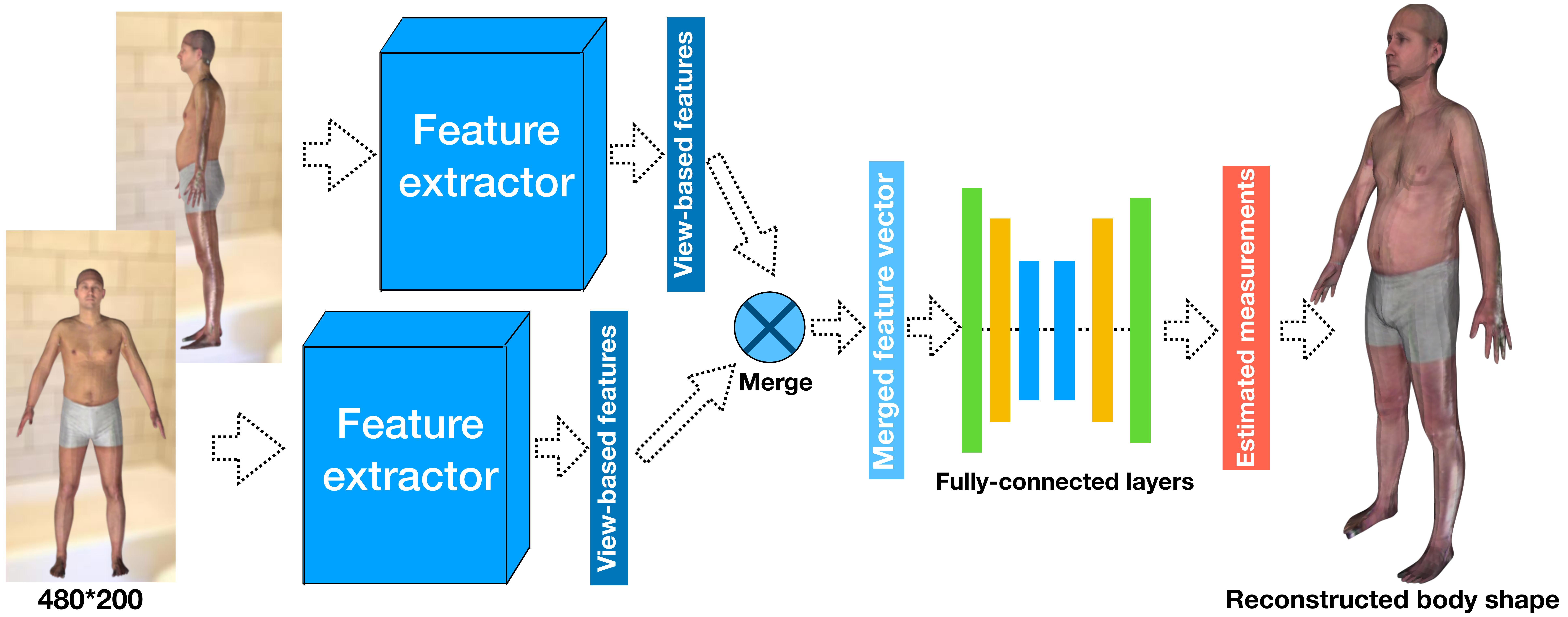}
   \caption{The proposed Anthropometric Measurements Network.}
    \label{fig:network}
\end{figure}

To tackle the task of estimating accurate anthropometric measurements from silhouettes or RGB images, we introduce a deep neural network architecture (Figure~\ref{fig:network}).
Unlike the previous works~\cite{dibra2016shape, dibra2016hs, Dibra-2017-cvpr, Bogo:ECCV:2016} whose primary task is body shape reconstruction, our network aims at learning a mapping from shape descriptors to anthropometric measurements. 
Our network consists of two components: 5 convolutional layers to compute
deep features for RGB or silhouette input and 6 fully-connected layers
to map the deep features to anthropometric body measurements. The network
can be trained with multiple inputs, 
but only two (frontal + side) were included to the experiments.
The subject height and virtual camera calibration parameters 
were used to scale and center the subject into an image of resolution $480\times200$. 
There is no weight sharing between the inputs to allow network to learn
a view specific features. For multiple inputs, a merge 
layer is applied to correlate the multiple view features before
the regression layer.
%
\section{Experiments}
\label{sec:experiments}
\noindent{\bf{Data Preparation}} 
We run experiments on two different datasets, XXX-fits and CAESAR-fits~\cite{yang2014semantic}.
The training and test samples are split equally for each dataset.
In order to generate a large number of training samples for network training,
the PCA-based statistical shape model constructed from the training samples is used to generate 10K training examples of various shapes. 

For each body part, the PCA is
applied separately and the first four principal components covering about $92\%$ of
shape variance are selected for learning the linear mapping to corresponding body measurements.

%
\noindent{\bf{Network Training \& Evaluation}} The proposed network learns
mapping from RGB images to 34 anthropometric measurements. The network is
trained with the Adadelta optimizer using the learning rate $10^{-4}$.
The network uses the standard MSE loss and is trained 100 epochs.

%
%
\subsection{Results}
\begin{figure}[t]
\centering
\includegraphics[width=\linewidth]{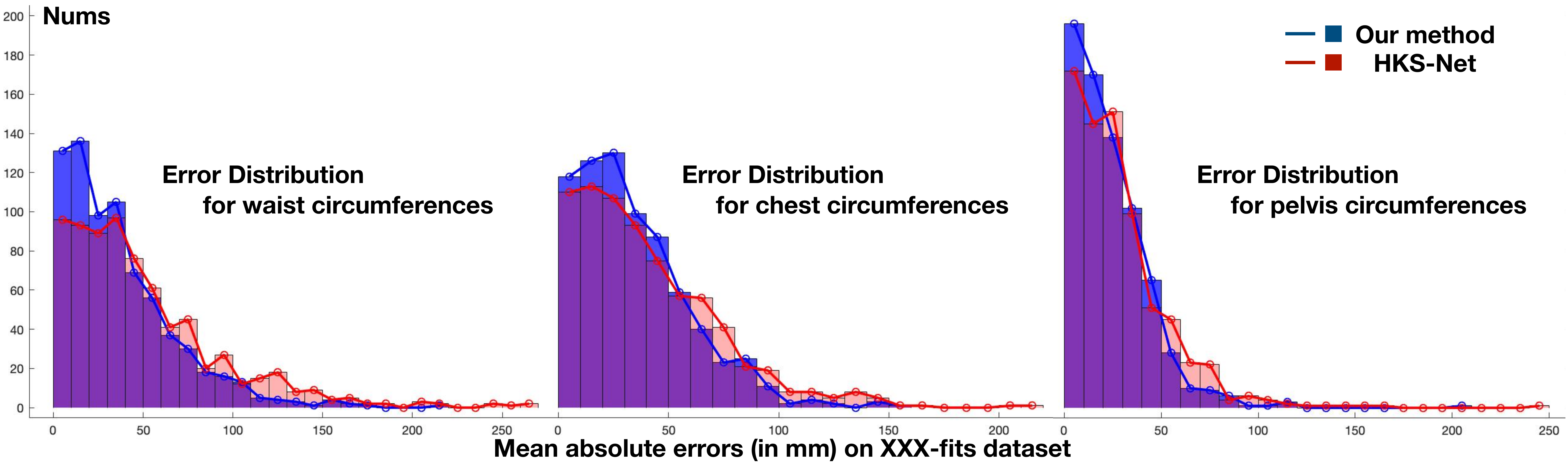}
\caption{Analysis of measurement errors for our method and HKS-Net on XXX-fits dataset.
Left: the error distribution for waist circumference; 
Middle: the error distribution for chest circumference;
Right: the error distribution for pelvis circumference.}
\label{fig:errorDist}
\end{figure}

{\bf{Quantitative Experiments}}
For comparison, the state-of-the-art methods,
HKS-Net~\cite{Dibra-2017-cvpr}, HMR~\cite{kanazawa2018end} and SMPLify~\cite{Bogo:ECCV:2016} are trained with the same data and compared
to the proposed network. HKS-Net uses the UF-US-2 architecture and was trained with RGB images. 
HMR~\cite{kanazawa2018end} and SMPLify~\cite{Bogo:ECCV:2016} use only the frontal RGB image. 
For SMPLify~\cite{Bogo:ECCV:2016} the estimated locations of joints by DeepCut~\cite{pishchulin2016deepcut} are provided and
the original models of \cite{Bogo:ECCV:2016} and \cite{kanazawa2018end} were used.
The mean measurement errors on reconstructed meshes are reported in Table~\ref{tab:singleTestOnNOMOfitsRGB} \&~\ref{tab:singleTestOnCAESARfitsRGB} and illustrations of the results are provided in the supplementary material.
Our method achieves competitive performance compared to the state-of-the-arts works on both two dataset. 
Our method shows significantly better performances on the upper torso (chest, waist and pelvis). 
The error distributions over these measurements for our method and HKS-Net on XXX-fits dataset are plotted in Figure~\ref{fig:errorDist}.



\begin{figure*}[t]
    \centering
    \includegraphics[width=0.85\linewidth]{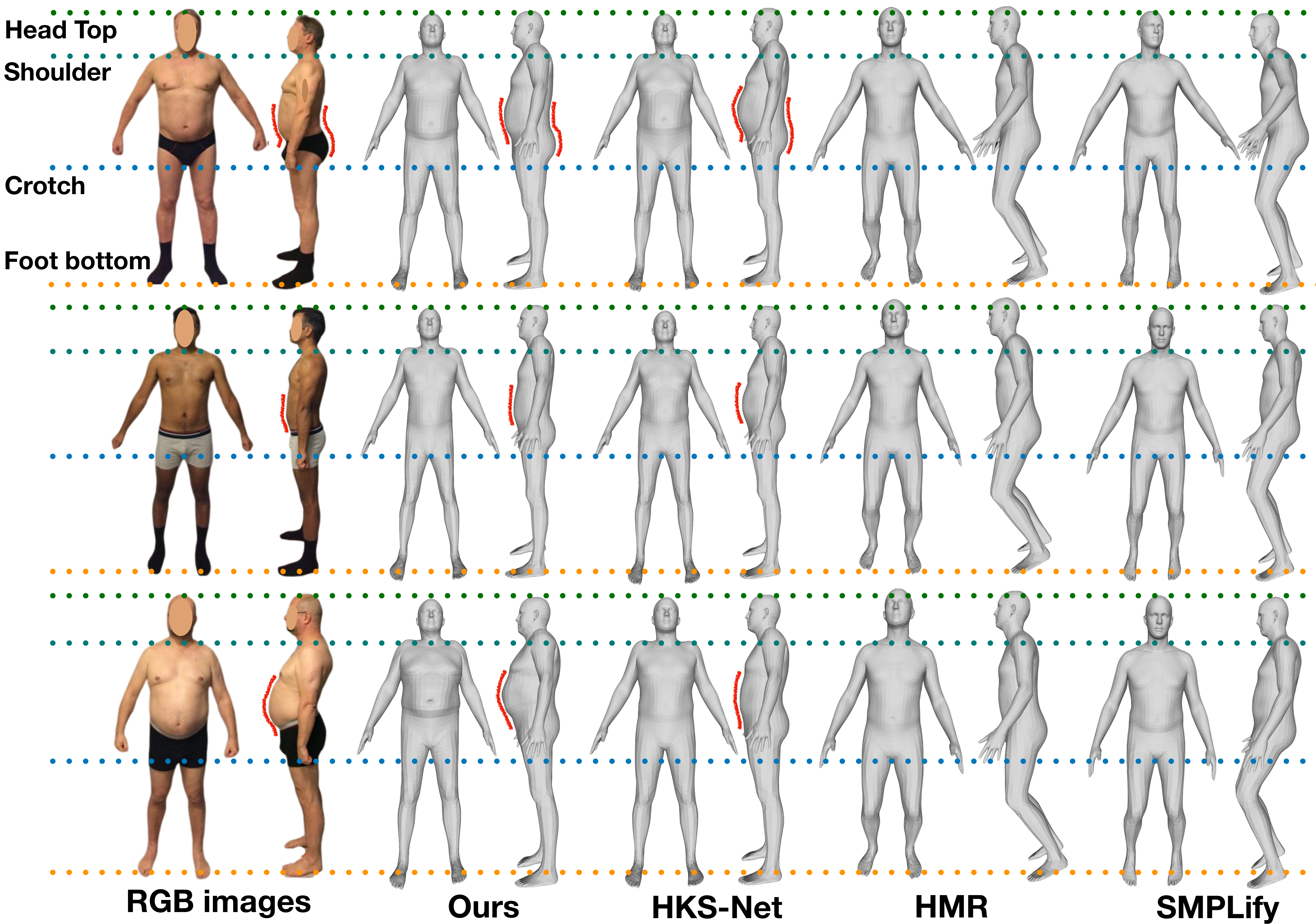}
   \caption{The Qualitative Experiments on XXX-real-200 dataset. 
   {\bf{Dash lines}} mark four meaningful locations: head top, shoulder, crotch point, and foot bottom.
   {\bf{Red lines}} mark the local areas where these methods particularly differ from each other. 
   {\bf{Bottom row}} shows the failure case due to the wrong estimations on the pelvis and upper torso.}
    \label{fig:qualitativeExp}
\end{figure*}

\begin{table}[h!]
\centering
 \begin{tabularx}{0.95\linewidth}
 {l| 
 >{\centering}>{\hsize=.03\textwidth}X
 >{\centering}>{\hsize=.03\textwidth}X
 >{\centering}>{\hsize=.03\textwidth}X
 >{\centering\arraybackslash}>{\hsize=.045\textwidth}X}
 \toprule
{\small{Measurements}} & Ours & {\small{HKS}}& {\small{HMR}} & {\small{SMPLify}} \\
 \hline
 {\small{a. Head Circumference}}    & 24.9          & {\bf{24.3}}  & 25.2 & ~33.0 \\ 
 {\small{b. Neck Circumference}}    & {\bf{14.5}}   & 15.8         & 25.3 & ~22.4 \\ 
 {\small{c. Shoulder-crotch Len.}}  & 14.8          & {\bf{13.2}}  & 25.7 & ~63.4 \\
 {\small{d. Chest Circumference}}   & {\bf{34.4}}   & 40.8         & 92.7 & ~67.3 \\
 {\small{e. Waist Circumference}}   & {\bf{36.7}}   & 50.3         & 88.7 & ~74.8 \\
 {\small{f. Pelvis Circumference}}  & {\bf{23.9}}   & 28.4         & 56.2 & ~89.3 \\
 {\small{g. Wrist Circumference}}   & ~7.9.         & {\bf{~7.3}}  & 11.0 & ~~9.9 \\
 {\small{h. Bicep Circumference}}   & {\bf{13.5}}   & 15.1         & 37.7 & ~26.4 \\
 {\small{i. Forearm Circumference}} & {\bf{~9.3}}   & ~9.7         & 16.1 & ~20.6 \\
 {\small{j. Arm length}}            & ~7.5          & {\bf{~5.9}}  & 24.5 & 138.0 \\
 {\small{k. Inside Leg length}}     & 12.9          & {\bf{10.0}}  & 36.3 & 229.6 \\
 {\small{l. Thigh Circumference}}   & {\bf{19.6}}   & 21.6         & 36.4 & ~44.8 \\
 {\small{m. Calf Circumference}}    & {\bf{11.2}}   & 12.7         & 19.7 & ~22.5 \\
 {\small{n. Ankle Circumference}}   & {\bf{~7.5}}   & {\bf{~7.5}}  & 10.7 & ~23.2 \\
 {\small{o. Overall height}}        & 29.8.         & {\bf{23.2}}  & 92.9 & 419.5 \\
 {\small{p. Shoulder breadth}}      & ~8.6          & {\bf{~7.9}}  & 19.9 & ~68.4 \\
 \bottomrule
 \end{tabularx}
 \caption{Experiments on RGB images of the XXX-fits dataset.
Mean absolute errors (MAE) of body measurements are reported in millimeters.
{\bf{Ours}} network predicts the body measurements, 
{\bf{HKS-Net}}~\cite{Dibra-2017-cvpr} adopts the UF-US-2 architecture;
{\bf{HMR}}~\cite{kanazawa2018end} and {\bf{SMPLify}}~\cite{Bogo:ECCV:2016} take only the frontal views.
SMPLify requires estimated joints from DeepCut~\cite{pishchulin2016deepcut}.
}
 \label{tab:singleTestOnNOMOfitsRGB}
\end{table}
%
\begin{table}[h!]
\centering
\begin{tabularx}{0.95\linewidth}
 {l| 
 >{\centering}>{\hsize=.03\textwidth}X
 >{\centering}>{\hsize=.03\textwidth}X
 >{\centering}>{\hsize=.03\textwidth}X
 >{\centering\arraybackslash}>{\hsize=.045\textwidth}X}
 \toprule
{\small{Measurements}} & Ours & {\small{HKS}}& {\small{HMR}} & {\small{SMPLify}} \\
 \hline
 {\small{a. Head Circumference}}    & 11.6         & {\bf{10.8}}   & 16.3 & ~28.1 \\ 
 {\small{b. Neck Circumference}}    & {\bf{12.3}}  & 13.1          & 27.2 & ~24.4 \\  
 {\small{c. Shoulder-crotch Len.}}  & 13.9         & {\bf{13.4}}   & 28.6 & ~57.8 \\ 
 {\small{d. Chest Circumference}}   & {\bf{26.1}}  & 28.3          & 68.3 & ~74.5 \\ 
 {\small{e. Waist Circumference}}   & {\bf{28.7}}  & 38.6          & 85.3 & ~72.8 \\ 
 {\small{f. Pelvis Circumference}}  & {\bf{22.6}}  & 26.0          & 62.8 & ~99.1 \\ 
 {\small{g. Wrist Circumference}}   & ~6.9         & {\bf{~6.5}}   & 14.3 & ~11.9 \\ 
 {\small{h. Bicep Circumference}}   & {\bf{13.0}}  & 13.4          & 35.6 & ~28.4 \\ 
 {\small{i. Forearm Circumference}} & {\bf{~7.8}}  & ~8.0          & 16.7 & ~25.9 \\ 
 {\small{j. Arm length}}            & ~9.5         & {\bf{~6.9}}   & 45.3 & 150.2 \\ 
 {\small{k. Inside Leg length}}     & 11.2         & {\bf{~8.7}}   & 37.2 & 219.1 \\ 
 {\small{l. Thigh Circumference}}   & {\bf{18.2}}  & 18.5          & 39.3 & ~51.3 \\ 
 {\small{m. Calf Circumference}}    & {\bf{11.7}}  & 11.8          & 21.4 & ~28.4 \\ 
 {\small{n. Ankle Circumference}}   & {\bf{~7.8}}  & ~7.9          & 13.6 & ~28.8 \\ 
 {\small{o. Overall height}}        & 20.1.        & {\bf{11.8}}   & 96.5 & 398.5 \\ 
 {\small{p. Shoulder breadth}}      & {\bf{~7.6}}  & ~7.7          & 21.8 & ~51.9 \\ 
 \bottomrule
 \end{tabularx}
 \caption{Experiments on RGB images of {\bf{CAESAR-fits}} dataset. 
MAEs of body measurements are reported in millimeters.}
 \label{tab:singleTestOnCAESARfitsRGB}
\end{table}
{\bf{Qualitative Experiments}} We evaluate related methods on the XXX-real-200 dataset.
Visualizations of some estimated body shapes are shown in Figure~\ref{fig:qualitativeExp}. 
Our reconstructions (the second column) restore finer local details, as compared to previous works~\cite{kanazawa2018end, Dibra-2017-cvpr, Bogo:ECCV:2016}.
Our method can be directly applied on the RGB images rather than converting images into binary silhouettes and does not require additional information, e.g. the estimated joints.

The bottom row in Figure~\ref{fig:qualitativeExp} shows the failure case due to wrong estimation on the lengths of upper torso and pelvis,
which leads to the unnatural ratio of upper body.
Interestingly similar mistakes happen in other methods.

{\bf{Different background images}} 
Considering about the effects brought by the background images, we evaluated the proposed network on the XXX-fits and CAESAR-fits datasets of which images are rendered with 4 types of background images: clear images, blurry images, random noisy images, and pure black images.
The mean measurement errors are illustrated in Table~\ref{tab:singleTestRGB_VS_Silh} and Figure~\ref{fig:differentBG}, and illustrations of the results are in the supplementary material.
Our network shows the robustness to complicated background images.
Binary silhouettes inputs provide the best results and the performance drops when the background getting complicated.
Background images, lighting, and cloths do bring negative effects and methods for background removal would promote the performance of anthropometric measurements estimation.
However, due to the imperfection of silhouette extraction algorithms,
it become difficult to obtain such perfect silhouettes.

\begin{figure}[t]
\centering
\includegraphics[width=0.9\linewidth]{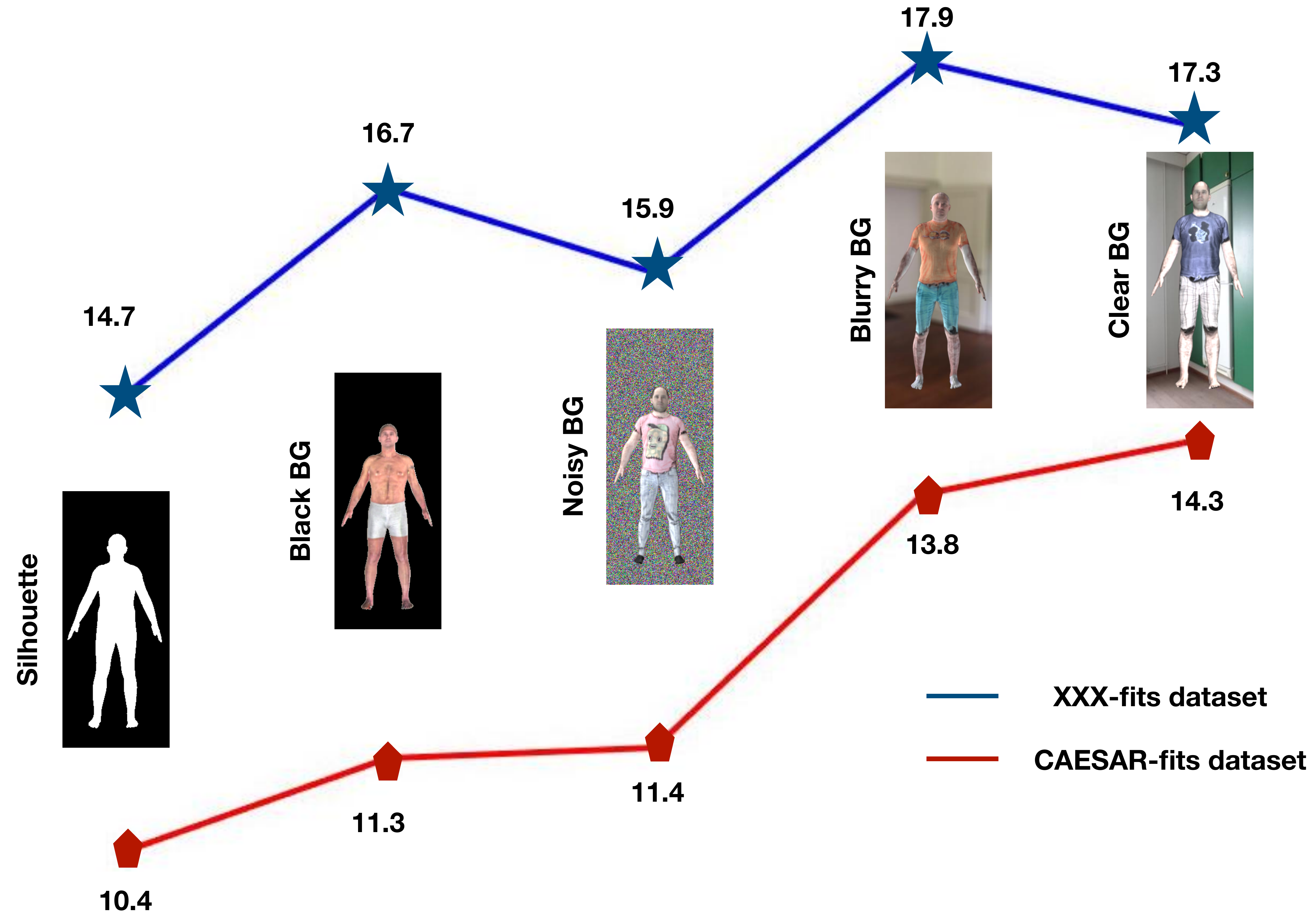}
\caption{Our proposed method is evaluated on XXX-fits and CAESAR-fits datasets with different background images .
{\bf{Mean errors}} of all 16 measurements (in mm) are plotted.}
\label{fig:differentBG}
\end{figure}
\begin{figure}[t]
\centering
\includegraphics[width=0.95\linewidth]{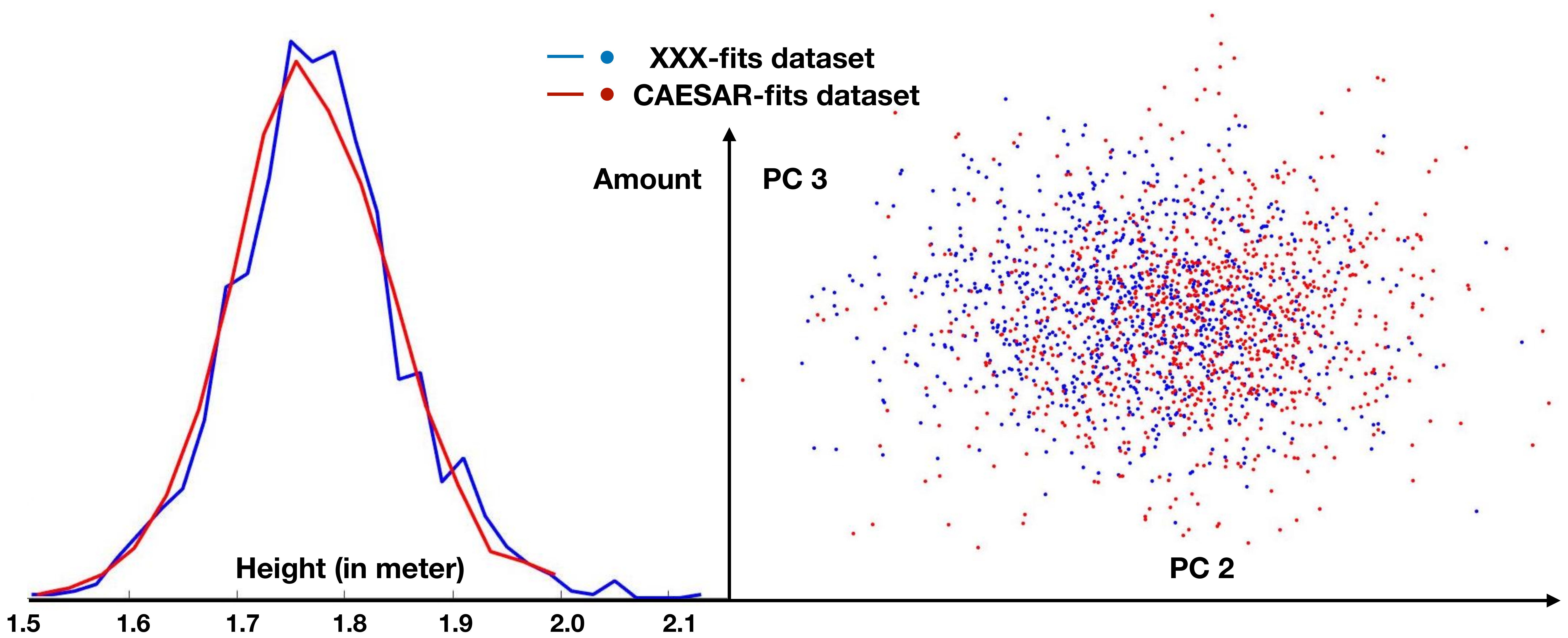}
\caption{Shape spaces of XXX-fits and CAESAR-fits datasets.
Left: the distributions of height data. Right: the first 2 PCA coefficients of shapes.}
\label{fig:shapeSpace}
\end{figure}
{\bf{Evaluation on Reconstructed bodies.}}
In our work, body shapes are recovered from estimated anthropometric measurements with the help of a part-based shape model (Sec~\ref{sec:semanticmapping}).
To illustrate the advantage of proposed part-based shape model, we train another network predicting totally 68 PCA coefficients for all parts.
The results of mean measurement errors on the reconstructed body surfaces are illustrated in Table~\ref{tab:PCA_vs_Meas_RGB}.
As shown, the linear mapping method restores the bodies in good qualities without losing local details compared to the network estimating PCA coefficients.

{\bf{Analysis of shape datasets.}}
To visualize high dimensional PCA shape spaces of the XXX-fits and CAESAR-fits datasets, we merge two datasets and perform PCA on these body meshes then select the first 10 PCA coefficients as the feature vectors and standardize them to have zero mean and unit variance.
To supply a lower-dimensional picture, we select the first 3 principle coefficients as the coordinates of body shapes.
Since the first principle component encodes the height information, we plot the distributions of height data from two datasets in Figure~\ref{fig:shapeSpace}.
Then the second and third principal coefficients are tread as the 2D coordinates.
Two datasets capture different shape variances and our proposed XXX-fits dataset contributes the considerable body shapes for related datasets and works.

{\bf{Discussion and Future works}}
A limitation to our method is that the body shape is reconstructed from 34 measurements covering the whole body. 
One challenge task is how to recover body shape from fewer (less than 34) measurements.
The correlation among anthropometric measurements would to be explored in future work.

Another one of future work is to consider how to narrow the gap between the self-defined measurements and tape measurements in related applications.
The gaps among different kinds of measurements are noticeable:
the self-defined body measurements of XXX-fits dataset (Sec~\ref{sec:measurementdefinition}), the TC2 measurements of XXX-scans dataset from the TC2 system and tape measurements of XXX-real-200 dataset.
For real applications, tape measurements are the foremost target and necessary procedures are required for domain adaption. 
In our experiments a non-linear regressor was trained on XXX-real-200 dataset for domain transfer, however, it is still insufficient to meet strict industrial requirements.
Analysis of measurement errors is illustrated in the supplementary material.
More data and works on vision-based anthropometry are needed in future work.
\section{Conclusions}
\label{sec:conclusions}
We posed the task of anthropometric measurements estimation as regression by learning a mapping from 2D image clues to body measurements,
and reconstruct body shapes from predicted measurements.
The proposed method was evaluated on thousands of human bodies (XXX-fits and CAESAR-fits datasets) and 200 real subjects (XXX-real-200 dataset).
To the authors' best knowledge 
the proposed dataset is the first freely available dataset of real human body shapes along with the measurements.
Further more, we evaluated the proposed method with images in different backgrounds and showed its robustness to the influence of noise of backgrounds, lighting and cloths .
\begin{table}[h!]
\centering
 \begin{tabularx}{0.95\linewidth}
 {l| 
 >{\centering}>{\hsize=.035\textwidth}X  >{\centering}>{\hsize=.035\textwidth}>{\columncolor[gray]{0.85}}X| 
 >{\centering}>{\hsize=.035\textwidth}X >{\centering\arraybackslash}>{\hsize=.035\textwidth}>{\columncolor[gray]{0.85}}X}
 \toprule
\multirow{2}{*}{\small{Measurements \& Datasets}}  & \multicolumn{2}{c|}{{\small{XXX-fits}}} & \multicolumn{2}{c}{{\small{CAESAR-fits}}} \\
   & RGB   & Silh.  & RGB    & Silh. \\
 \hline
 {\small{a. Head Circumference}}    & 24.9  & 22.8   & 11.6   & 10.4  \\ 
 {\small{b. Neck Circumference}}    & 14.5  & 14.4   & 12.3   & 10.6  \\ 
 {\small{c. Shoulder-crotch Len.}}  & 14.8  & 12.4   & 13.9   & 12.9  \\
 {\small{d. Chest Circumference}}   & 34.4  & 22.2   & 26.1   & 15.3  \\
 {\small{e. Waist Circumference}}   & 36.7  & 32.9   & 28.7   & 15.0  \\
 {\small{f. Pelvis Circumference}}  & 23.9  & 23.8   & 22.6   & 17.0  \\
 {\small{g. Wrist Circumference}}   & ~7.9  & ~7.4   & ~6.9   & ~6.3  \\
 {\small{h. Bicep Circumference}}   & 13.5  & 10.6   & 13.0   & ~9.9  \\
 {\small{i. Forearm Circumference}} & ~9.3  & ~7.7   & ~7.8   & ~6.2  \\
 {\small{j. Arm length}}            & ~7.5  & ~5.4   & ~9.5   & ~6.4  \\
 {\small{k. Inside Leg length}}     & 12.9  & ~8.5   & 11.2   & ~7.1  \\
 {\small{l. Thigh Circumference}}   & 19.6  & 18.9.  & 18.2   & 12.8  \\
 {\small{m. Calf Circumference}}    & 11.2  & 11.7   & 11.7   & 10.6  \\
 {\small{n. Ankle Circumference}}   & ~7.5  & ~7.1   & ~7.8   & ~7.2  \\
 {\small{o. Overall height}}        & 29.8  & 21.9   & 20.1   & 12.4  \\
 {\small{p. Shoulder breadth}}      & ~8.6  & ~6.9   & ~7.6   & ~6.0  \\
 \bottomrule
 \end{tabularx}
 \caption{The results of our proposed method with RGB images and binary silhouettes on XXX-fits and CAESAR-fits datasets.
Mean absolute errors of body measurements are reported in millimeters.
}
 \label{tab:singleTestRGB_VS_Silh}
\end{table}
%
\begin{table}[h!]
\centering
 \begin{tabularx}{0.97\linewidth}
 {c| 
 >{\centering}>{\hsize=.04\textwidth}X  >{\centering}>{\hsize=.038\textwidth}X >{\centering}>{\hsize=.038\textwidth}>{\columncolor[gray]{0.85}}X|
 >{\centering}>{\hsize=.04\textwidth}X  >{\centering}>{\hsize=.038\textwidth}X >{\centering\arraybackslash}>{\hsize=.038\textwidth}>{\columncolor[gray]{0.85}}X}
 \toprule
{\small{Datasets}} & \multicolumn{3}{c|}{{\small{XXX-fits}}} & \multicolumn{3}{c}{{\small{CAESAR-fits}}}\\
\hline
{\small{Meas.}}   & {\tiny{Part-PCA}} & {\tiny{Meas-1}} & {\tiny{Meas-2}}  & {\tiny{Part-PCA}} & {\tiny{Meas-1}} & {\tiny{Meas-2}}\\
 \hline
 {\small{a.}}   & 23.1  & 23.0  & 24.9   & 10.7   & 11.3 & 11.6 \\ 
 {\small{b.}}   & 15.4  & 14.5  & 14.5   & 12.1   & 12.1 & 12.3 \\ 
 {\small{c.}}   & 18.5  & 18.7  & 14.8   & 16.1   & 16.4 & 13.9 \\
 {\small{d.}}   & 33.5  & 34.4  & 34.4   & 26.9   & 26.0 & 26.1 \\
 {\small{e.}}   & 45.0  & 36.5  & 36.7   & 32.2   & 28.5 & 28.7 \\
 {\small{f.}}   & 26.6  & 23.9  & 23.9   & 23.4   & 22.8 & 22.6 \\
 {\small{g.}}   & ~7.1  & ~7.2  & ~7.9   & ~6.5   & ~6.8 & ~6.9 \\
 {\small{h.}}   & 13.0  & 13.6  & 13.5   & 12.2   & 13.0 & 13.0 \\
 {\small{i.}}   & ~8.9  & ~9.1  & ~9.3   & ~7.6   & ~8.0 & ~7.8 \\
 {\small{j.}}   & ~6.0  & ~7.6  & ~7.5   & ~7.0   & ~9.5 & ~9.5 \\
 {\small{k.}}   & 10.2  & 13.0  & 12.9   & ~8.2   & 11.2 & 11.2 \\
 {\small{l.}}   & 20.0  & 19.7  & 19.6   & 16.9   & 18.2 & 18.2 \\
 {\small{m.}}   & 11.8  & 11.2  & 11.2   & 11.4   & 11.7 & 11.7 \\
 {\small{n.}}   & ~7.5  & ~7.5  & ~7.5   & ~7.7   & ~7.8 & ~7.8 \\
 {\small{o.}}   & 25.7  & 32.1  & 29.8   & 17.5   & 21.9 & 20.1 \\
 {\small{p.}}   & ~9.0  & ~9.1  & ~8.6   & ~7.3   & ~7.7 & ~7.6 \\
 \bottomrule
 \end{tabularx}
 \caption{Evaluation on the accuracy of reconstructed body shapes.
Mean absolute errors (MAE) of body measurements on the {\bf{reconstructed body surfaces}} are reported in millimeters.
{\bf{Part-PCA}} denotes the body shapes are reconstructed by the PCA coefficients for body parts (Sec~\ref{sec:subpcashapemodel});
{\bf{Meas-1}} denotes that the network directly predicts the body measurements and body shapes are reconstructed through the linear mapping  (Sec~\ref{sec:semanticmapping});
{\bf{Meas-2}} denotes the MAE of the predicted measurements.}
 \label{tab:PCA_vs_Meas_RGB}
\end{table}

{\small
\bibliographystyle{ieee_fullname}
\bibliography{human3d}
}

\end{document}